\documentclass[doublecolumn,oneside,letter]{IEEEconf}
\usepackage{amsmath}
\usepackage{amssymb}
\usepackage{graphicx}
\usepackage{subfigure}
\usepackage{verbatim}
\usepackage[thinlines,thiklines]{easybmat}
\usepackage{latexsym}
\usepackage[dvipsnames]{xcolor}
\usepackage{cite}
\usepackage{stmaryrd}
\usepackage{multirow}
\usepackage{textcomp}
\usepackage[ruled]{algorithm2e}
\usepackage{diagbox}
\usepackage{array}
\usepackage{bm}
\usepackage{color,soul}
\sethlcolor{yellow} 
\usepackage[left=0.75in,top=0.75in,right=0.75in,bottom=0.75in]{geometry}
\IEEEoverridecommandlockouts
\def\s{\mathop{\rm s}\nolimits}
\def\c{\mathop{\rm c}\nolimits}
\newcommand{\bs}[1]{\ensuremath{{\boldsymbol{#1}}}}
\newcommand{\mathbi}[1]{\ensuremath{{\boldsymbol{#1}}}}

\IEEEoverridecommandlockouts 
\begin{document}

\title{\Large \bf Foot Shape-Dependent Resistive Force Model for Bipedal Walkers \\ on Granular Terrains~\thanks{The work was supported in part by the US NSF under award CMMI-2222880.}}

\author{Xunjie Chen\thanks{X. Chen, A. Anikode and J. Yi are with the Department of Mechanical and Aerospace Engineering, Rutgers University, Piscataway, NJ 08854 USA (email: xc337@rutgers.edu, aaa387@scarletmail.rutgers.edu, jgyi@rutgers.edu).}, Aditya Anikode, Jingang Yi, and Tao Liu\thanks{T. Liu is with the State Key Lab of Fluid Power Transmission and Control and the School of Mechanical Engineering, Zhejiang University, Hangzhou, Zhejiang 310027, China (email: {liutao@zju.edu.cn}).}}

\maketitle
\thispagestyle{empty}
\pagestyle{empty}

\begin{abstract}
Legged robots have demonstrated high efficiency and effectiveness in unstructured and dynamic environments. However, it is still challenging for legged robots to achieve rapid and efficient locomotion on deformable, yielding substrates, such as granular terrains. We present an enhanced resistive force model for bipedal walkers on soft granular terrains by introducing effective intrusion depth correction. The enhanced force model captures fundamental kinetic results considering the robot foot shape, walking gait speed variation, and energy expense. The model is validated by extensive foot intrusion experiments with a bipedal robot. The results confirm the model accuracy on the given type of granular terrains. The model can be further integrated with the motion control of bipedal robotic walkers.
\end{abstract}

\section{Introduction}

Recent advances in legged robotics have demonstrated high efficiency and effectiveness in unstructured and dynamic environments. It is however still challenging for legged robots to achieve rapid and efficient operation on deformable, yielding terrains such as sand~\cite{AguilarRPP2016,GodonFRAI2023}. Foot-terrain modeling plays an important role for developing safe and efficient locomotion control on yielding substrates. Wheel-terrain interactions (i.e., terramechanics) have been studied for vehicles to navigate on off-road, soils or sand surface for decades~\cite{Wong2001b,AgarwalSA2021}. For legged robots, various foot-terrain interaction models were proposed from physical principles to capture the interaction forces (e.g.,~\cite{DingIJRR2013}). In~\cite{LiScience2013}, a resistive force theory (RFT) model was presented for arbitrarily-shape legs that move freely in granular materials with different depths and orientations. Xu {\em et al}.~\cite{XuJTM2015} proposed a hybrid force model that integrated the RFT with a failure-based model to predict the thrust and support forces for various-shape legs in granular materials. The work in~\cite{AgarwalSA2021} proposed an reduced-order RFT model by considering the dynamic effects of arbitrarily shaped intruders in granular materials. The research work in~\cite{treers2021granular} further extended the dynamic RFT model to three-dimensional directions. Huang {\em et al.}~\cite{HuangRAL2022} applied the RFT model to a screw-propelled wheeled robot optimization.

Computational approaches such as the material point method were developed to compute the drag and lifting support forces in granular materials~\cite{AgarwalPhD2022}. These computational models however are not feasible for real-time applications. Experimental approaches provide promising results to study the foot-terrain interactions of legged robots. In~\cite{DingIJRR2013}, a cylindrical foot was used to validate the force models on various foot and terrain materials. In~\cite{VanderICRA2022}, a flat-shape robotic foot was used to test a new terramachanic model on three types of foot materials on several soil types. Machine learning-based methods were also developed to detect and classify the planetary soil types by using legged robotic feet~\cite{KolvenRAL2019}. The experiments in~\cite{DingIJRR2013,VanderICRA2022,KolvenRAL2019,GodonRAL2022} only considered the regular shape foot and the leg motion was only in one-dimensional stepping motion.

There are limited studies on human walking locomotion on granular terrains. In~\cite{YangACCESS2019}, foot-terrain interactions were analyzed for human walking on sand and snow terrains. The stepping force and applied ankle torques were among the interests. The work in~\cite{SvennHMS2019} compared the human walking gaits on sand with those on the firm ground. Kinematics such as lower-limb joint angles and center-of-mass position were found significantly different for walking on sand terrain with on the firm ground. Energy expense for human walking on sand terrain was reported in~\cite{LejeuneJEB1998}. No kinetic results are reported for human walking on granular terrains. Most of the studies of legged robots on granular materials include the circular-leg RHex (e.g.,~\cite{ZhangIJRR2013}), single-leg hopping robot (e.g.,~\cite{GartBB2021}), and multi-leg robots (e.g.,~\cite{DingRAL2022}), etc., and few on bipedal robotic walkers~\cite{XiongIROS2017}. In~\cite{ChenMECC2023}, foot shape was analyzed and optimized for high energy efficiency for bipedal walkers and only simulation was conducted.
 
In this paper, we extend the work in \cite{ChenMECC2023} for an updated RFT model for bipedal walking locomotion. One goal of this work is to investigate the effect of the robot foot shape and gait on bipedal walking locomotion on granular substrates. We first consider human walking gaits and locomotion on yielding substrates to reconstruct the foot intrusion gait profile for bipedal walkers. Unlike the previous work in which only two-dimensional (2D) force was considered, three-dimensional (3D) formulation is added with the intrusion orientation. Meanwhile, the inertial effect of substrates given a relative intrusion velocity is also considered for walking gait. Compared with the previous RFT model, the effective intrusion depth is defined and a structural correction term is introduced to improve the prediction accuracy and computational time. Based on the force model and kinetic calculation, we analyze different factors such as gait speed and foot shape for foot-terrain interactions. Experimental studies are conducted to validate and demonstrate the proposed force model. The main contribution of this work lies in the new enhanced RFT method for bipedal robotic locomotion on granular terrains and experimental evaluation considering foot shape and gait speed variations.

\section{Dynamic RFT Model for Bipedal Foot}
\label{Sec:model}

\subsection{3D RFT Configuration}

\begin{figure}[ht!]
\vspace{-1mm}
	\centering
	\subfigure[]{
		\label{fig:3DRFT} \includegraphics[width=2.6in]{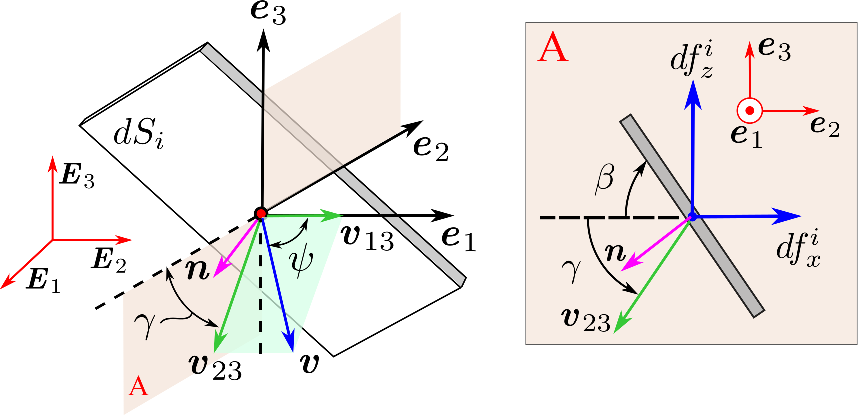}}
	\subfigure[]{
		\label{fig:deltaH_computation}
\includegraphics[width=2.2in]{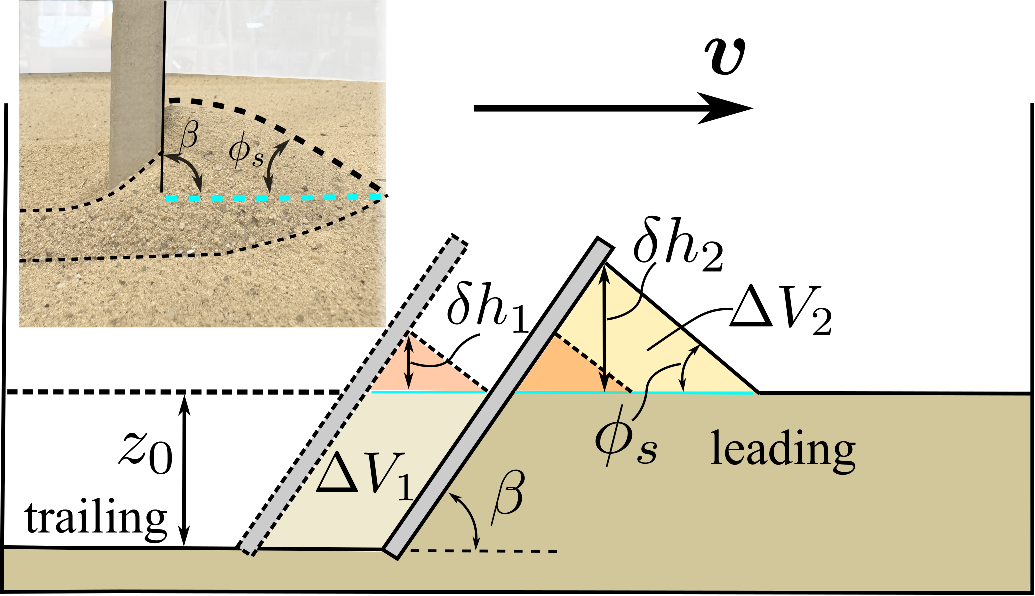}}
\vspace{-2mm}
	\caption{(a) The general 3D plate intrusion configuration of plate $dS_i$. (b) The schematic of horizontal intrusion with the height difference of the free surfaces in the leading and trailing zones.}
\vspace{-1mm}
\end{figure}

Because of the complex contour of robot feet, we formulate a 3D resistive force model by discretizing foot surface into finite individual flat intrusion plates~\cite{treers2021granular,HuangRAL2022}. Fig.~\ref{fig:3DRFT} illustrates an arbitrary infinitesimal plate with surface area $dS_i$ that intrudes in substrates with translational velocity $\bs{v}$ and (point-out) normal vector $\bs{n}$. The global and body-fixed frames are defined with unit vectors $\mathbi{E}_i$ and $\bs{e}_i$, respectively, $i=1,2,3$. We define $\mathbi{e}_3 = \mathbi{E}_3$, $\mathbi{e}_2$ is located in the plane determined by $\mathbi{n}$ and $\bs{e}_3$, and the remaining axis is determined as $\mathbi{e}_1=\mathbi{e}_2\times \mathbi{e}_3$.

The 3D plate intrusion angle configuration is defined as follows. The intrusion orientation angle $\beta \in [-\frac{\pi}{2},\frac{\pi}{2}]$ is defined as the angle between the plate and $\bs{e}_2$, and then the angle between the vector $\mathbi{n}$ and the $\mathbi{e}_2$-axis is $\beta-\frac{\pi}{2}$. The independent angle $\psi$ is defined as the angle between velocity $\boldsymbol{v}$ and $\mathbi{e}_1$. $\bs{v}_{13}$ and $\bs{v}_{23}$ are the velocity components along the $\mathbi{e}_1$-axis and in the plane $\mathbi{e}_2 \mathbi{e}_3$, respectively, namely, $\bs{v}_{13}=\|\bs{v}\|\cos{\psi}\mathbi{e}_1$ and $\bs{v}_{23}=\bs{v}-\bs{v}_{13}$. The intrusion angle $\gamma \in [-\frac{\pi}{2},\frac{\pi}{2}]$ is the angle between $\bs{v}_{23}$ and the $\mathbi{e}_2$-axis. By this frame, the plate motion is decomposed into two components, namely, one 2D in-plane motion perpendicular to the plate (plane $A$ in Fig.~\ref{fig:3DRFT}) and the other sliding motion along the plate surface, i.e., $\mathbi{e}_1$-direction.

For $dS_i$, the force $d\mathbi{F}_i$ caused by the flow resistance and yield stress of granular substrates is computed as
\begin{equation}\label{eqn:dF}
  d \mathbi{F}_i = f_{i,1}(\boldsymbol{v},\mathbi{e}_1) d\mathbi{F}_{i,1} + f_{i,23}(\boldsymbol{v},\mathbi{e}_1) d\mathbi{F}_{i,23},
\end{equation}
where $d\mathbi{F}_{i,1}$ and $d\mathbi{F}_{i,23}$ are the resistive force components along the $\mathbi{e}_1$ direction and projected in the plane $\mathbi{e}_2 \mathbi{e}_3$, respectively. Two dimensionless scaling factors $f_{i,1}(\boldsymbol{v},\mathbi{e}_1)$ and $f_{i,23}(\boldsymbol{v},\mathbi{e}_1)$ are introduced to weight two components and their values are obtained by experiments. The two factors are related to the plate intrusion orientation, namely, $\psi$ and $\gamma$. Intuitively, $f_{i,1}(\boldsymbol{v},\mathbi{e}_1)=1$ and $f_{i,23}(\boldsymbol{v},\mathbi{e}_1)=0$ when $\psi=0$. If $\psi = \frac{\pi}{2}$, that is the intrusion process is purely an in-plane motion of $\mathbi{e}_2 \mathbi{e}_3$, $f_{i,1}(\boldsymbol{v},\mathbi{e}_1)=0$ and $f_{i,23}(\boldsymbol{v},\mathbi{e}_1)=1$.

By the RFT method~\cite{LiScience2013}\cite{treers2021granular}, two force components in~\eqref{eqn:dF} are determined by
\begin{subequations}\label{eqn:2D_RFT_decomposition}
  \begin{align}
    d\mathbi{F}_{i,1} & =\alpha_y(\beta,\gamma)|z|H(-z)\mathbi{e}_1 dS_i, \\
    d\mathbi{F}_{i,23} & =\left[ \alpha_x(\beta,\gamma)\mathbi{e}_2+ \alpha_z(\beta,\gamma)\mathbi{e}_3 \right]|z|H(-z) dS_i,
  \end{align}
\end{subequations}
where $|z|$ is the intrusion depth in substrates, $H(x)$ is Heaviside function, that is, $H(x)=1$ when $x>0$; otherwise, $H(x)=0$. In~\eqref{eqn:2D_RFT_decomposition}, $\alpha_{j}(\beta,\gamma)$, $j=y,x,z$, reflects local stresses (per unit depth) in the corresponding directions of $\mathbi{e}_{i}$, $i=1,2,3$, respectively. The local stresses are characterized by small plate intrusion experiments. In this paper, we estimate $\alpha_y(\beta,\gamma)=\alpha_x(0,0)$ to reduce additional experimental measurement fitting for convenient adoption. It is reported that granular materials share similar local stress characteristics. Using a scaling factor $\zeta$, the local stress maps can be obtained as $\alpha_j = \zeta \alpha_{j0}, ~(j=x,z)$
where $\zeta \approx 0.8 \alpha_z(0,\pi/2)$ determined from the vertical penetration experiment. Fourier series are used to fit $\alpha_{j0}$ that describes generic granular material properties~\cite{LiScience2013}.

\subsection{Dynamic RFT with Intrusion Depth Correction}

Inspired by the work in~\cite{AgarwalSA2021,HuangRAL2022,schiebel2020mitigating}, for intrusion plate $dS_i$, additional force representing dynamic inertial effect is introduced as a rate-dependent form as
\begin{equation}\label{eqn:inertialForce}
  d\mathbi{F}_{i,v} = -\lambda_{v} \rho \|\langle \boldsymbol{v}, \boldsymbol{n}\rangle\|^2 \hat{\boldsymbol{v}} dS_i, ~\hat{\boldsymbol{v}}=\frac{\boldsymbol{v}}{\|\boldsymbol{v}\|},
\end{equation}
where $\rho$ is the effective granular substrate density and scaling factor $\lambda_{v}$ is determined by experiments. In~\eqref{eqn:inertialForce}, the negative sign indicates the force is against the velocity direction.

The dynamic inertial term in~\eqref{eqn:inertialForce} is however insufficient to characterize the resistance force of the rapid flow. An additional dynamic term is introduced to describe pressure reduction caused by the height difference of the free surfaces in the leading and trailing zones. Fig.~\ref{fig:deltaH_computation} illustrates the height difference during the horizontal intrusion of the plate in granular material. Because of the intrusion, substrates are pushed by the intruder in the leading front face, forming a partial cone and making the height of the free surface in the leading zone higher than that in the trailing zone. The height difference is denoted as $z_0 + \delta h$, where $z_0$ is the intrusion depth in the trailing zone and $\delta h$ is the height with respect to the free surface in the front zone. The volume of cone grows as the intrusion process until it cannot sustain the weight due to the friction limitation. The base angle of the substrate cone, denoted by $\phi_s$, is related to the internal friction properties of the granular media rather than the shape of the intruder plate.

The steady height difference $z_0 + \delta h$ inevitably brings the pressure gradient of two sides of the intruder and therefore, generates an additional resistance force source. To simplify the calculation, we assume that (1) the intrusion is an in-plane motion with constant velocity $\boldsymbol{v}$; (2) The steady condition is considered and the intrusion angle $\beta$ and cone base angle $\phi_s$ maintain constant during the horizontal intrusion process. As shown in Fig.~\ref{fig:deltaH_computation}, the cone volume is calculated as
\begin{equation}
  V_{c,k} = \frac{1}{2}\left(\frac{1}{\tan{\beta}} + \frac{1}{\tan{\phi_s}} \right)\delta h_{k}^2=\frac{1}{2g(\beta,\phi_s)}\delta h_{k}^2,
\end{equation}
where $g(\beta,\phi_s)=\left(\frac{1}{\tan{\beta}} + \frac{1}{\tan{\phi_s}} \right)^{-1}$. Then, the corresponding volume change is
\begin{equation}
  \Delta V_2 = V_{c,2}-V_{c,1} = \frac{1}{2g(\beta,\phi_s)}\left(\delta h_{2}^2-\delta h_{1}^2\right).
\end{equation}
The volume occupied by the intruder beneath the free surface during a small time interval is  $\Delta V_1 = \|\boldsymbol{v}\| z_0 \Delta t$. This portion of volume change is assumed to contribute to the volume increscent of the cone in the leading zone such that $\Delta V_2 = \Delta V_1$. We compute the change rate of $S_h=\delta h^2$
\begin{equation}
  \frac{\Delta S_h}{\Delta t} = 2 \|\boldsymbol{v}\| z_0 g(\beta,\phi_s).
\end{equation}
Therefore, the cone height is intuitively proportional to the velocity and intrusion depth, that is, $\delta h \propto \sqrt{\|\boldsymbol{v}\| z_0 g(\beta,\phi_s)}$.

Introducing this dynamic structural correction, we modify resistance force calculation in the $\mathbi{e}_2\mathbi{e}_3$ plane of \eqref{eqn:2D_RFT_decomposition} considering the pressure gradient for each intruder plate,
\begin{equation}\label{eqn:modified_dF23}
  d\tilde{\mathbi{F}}_{i,23} = \left[ \alpha_x(\beta,\gamma)\mathbi{e}_2+ \alpha_z(\beta,\gamma)\mathbi{e}_3 \right]|\tilde{z}|H(-\tilde{z}) dS_i.
\end{equation}
In \eqref{eqn:modified_dF23}, the effective intrusion depth is defined as
\begin{equation}\label{eqn:effectiveDepth}
  \tilde{z}= z_0 + \delta h = z_0 + \lambda_{h}\sqrt{\|\boldsymbol{v}\| z_0 g(\beta,\phi_s)},
\end{equation}
where $\lambda_{h}$ is the scaling factor that is obtained by horizontal penetration tests.

For 3D dynamic intrusion, considering~\eqref{eqn:dF},~\eqref{eqn:inertialForce} and~\eqref{eqn:modified_dF23}, the element resistive force is summarized as
\begin{equation}\label{eqn:modified_3D RFT}
    d \mathbi{F}_i = f_{i,1}(\boldsymbol{v},\mathbi{e}_1) d\mathbi{F}_{i,1} + f_{i,23}(\boldsymbol{v},\mathbi{e}_1) d\tilde{\mathbi{F}}_{i,23} + d\mathbi{F}_{i,v}.
\end{equation}
Total reaction force is then computed as $\mathbi{F}_{RFT} = \int_S d\mathbi{F}_i$.

\subsection{Foot Intrusion Motion of Bipedal Walking}
\label{Sec:opt}

\begin{figure}[h!]
\vspace{2mm}
  \centering
  \includegraphics[width=2.5in]{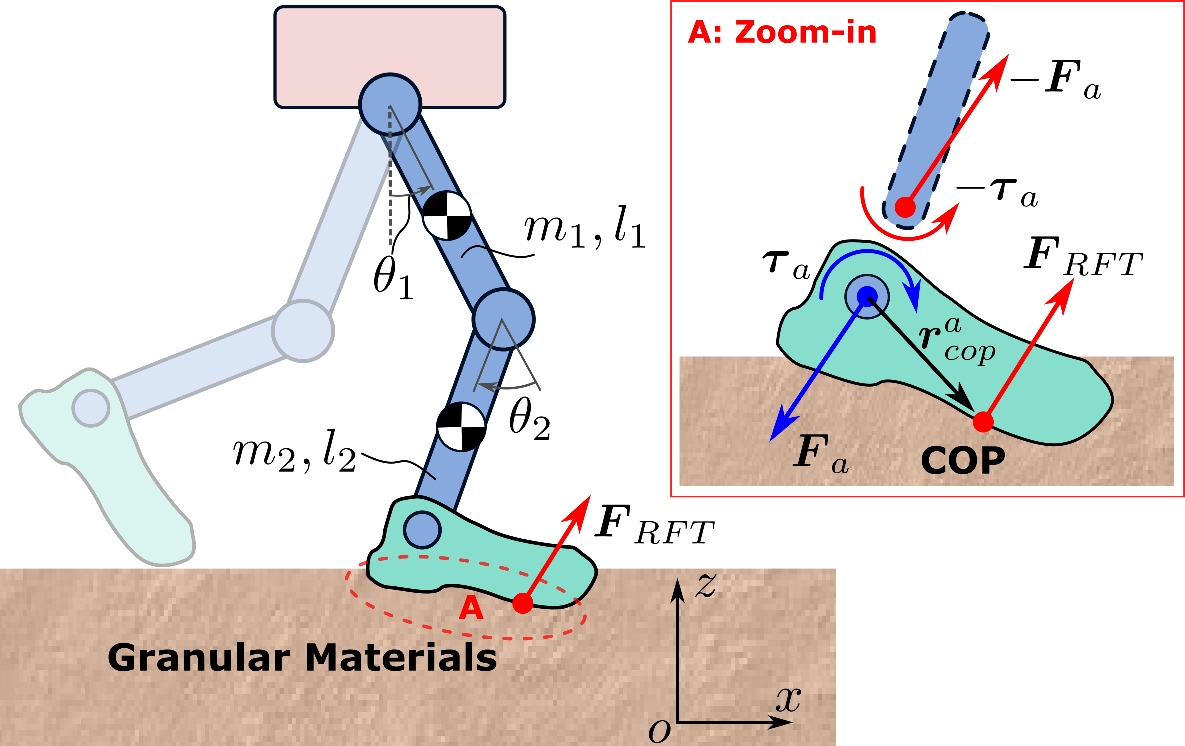}
  \caption{Bipedal robot walks on the granular terrain (single stance phase).}\label{fig:robotWalker}
\vspace{-2mm}
\end{figure}

Fig.~\ref{fig:robotWalker} illustrates the stance phase of the bipedal robot walking on the granular terrain. A double-link model is used to present the motion of the robotic walker in the sagittal plane. The hip (with respect to the vertical direction) and knee angles are defined as $\theta_1$ and $\theta_2$, respectively. The foot is lightweight and there is no actuator for the ankle joint. We focus on the external force distribution by contact along the foot contour. Therefore, it is assumed that no moment is applied to the feet from the terrain.

For the foot in the intrusion process, we denote the external force and torque at the ankle joint as $\mathbi{F}_a$ and $\tau_{a}$, respectively. We consider a “quasi-static" condition for the robot foot,
\begin{equation}\label{eqn:footBalance}
    \mathbi{F}_a + \mathbi{F}_{RFT} = \mathbi{0},~
     \mathbi{r}_{COP}^a \times\mathbi{F}_{RFT}=\tau_a,
\end{equation}
where $\mathbi{r}_{COP}^a$ is the position vector of the center of pressure (COP) of the foot with respect to the ankle joint, $\mathbi{r}_{COP}^a= \mathbi{r}_{COP}-\mathbi{r}_{a}$. It is straightforward to obtain the force distribution along the foot and therefore, we estimate the COP position,
\begin{equation}\label{eqn:COP}
  r_{COP,x} = \frac{\int z_i d\mathbi{F}_{i,x}(z)}{\int_S d\mathbi{F}_{i,x}}, ~r_{COP,z} = \frac{\int x_i d\mathbi{F}_{i,z}(x)}{\int_S d\mathbi{F}_{i,z}}.
\end{equation}
The corresponding power $P$ and work $W$ performed by joint actuators in a given stance interval $t_s$ can be calculated as $P = \boldsymbol{\tau} \cdot \dot{\mathbi{q}}$, $W = \int_{0}^{t_s} P dt$, where $\mathbi{q}=[\theta_1~\theta_2]^T$, and $\mathbi{\tau}=[\tau_1~\tau_2]^T$ is the joint torque vector that is determined by the inverse kinetics model with $\mathbi{F}_a$ and $\tau_{a}$.

The computation of the reaction force based on the dynamic RFT allows us to quantify contact resistance and obtain the distribution of resistive force on foot contour. This shape-dependent information can be used for optimization of bipedal robotic foot design~\cite{ChenMECC2023}. Furthermore, this force model can be extended and integrated with whole-body dynamics models (e.g.~\cite{chen2015robotic,trkov2019bipedal}) of bipedal walkers on soft terrains. Furthermore, it would benefit the robot balance sensing, control and strategies design to enhance walking stability (e.g.,~\cite{TrkovTASE2019,MihalecTMech2023}).

\section{Experiments}
\label{Sec:exp}

\begin{figure*}[h!]
\hspace{2mm}
\subfigure[]{
 \label{fig:experimentSetup:a}
  \includegraphics[height = 1.75in]{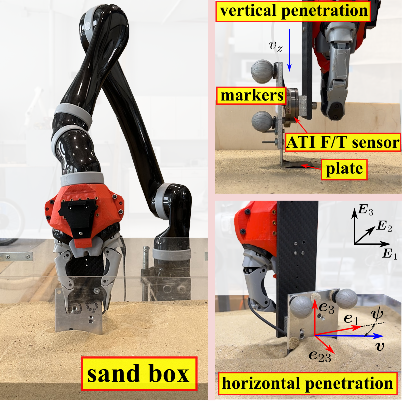}}
\subfigure[]{
	\label{fig:experimentSetup:b}
	\includegraphics[height = 1.75in]{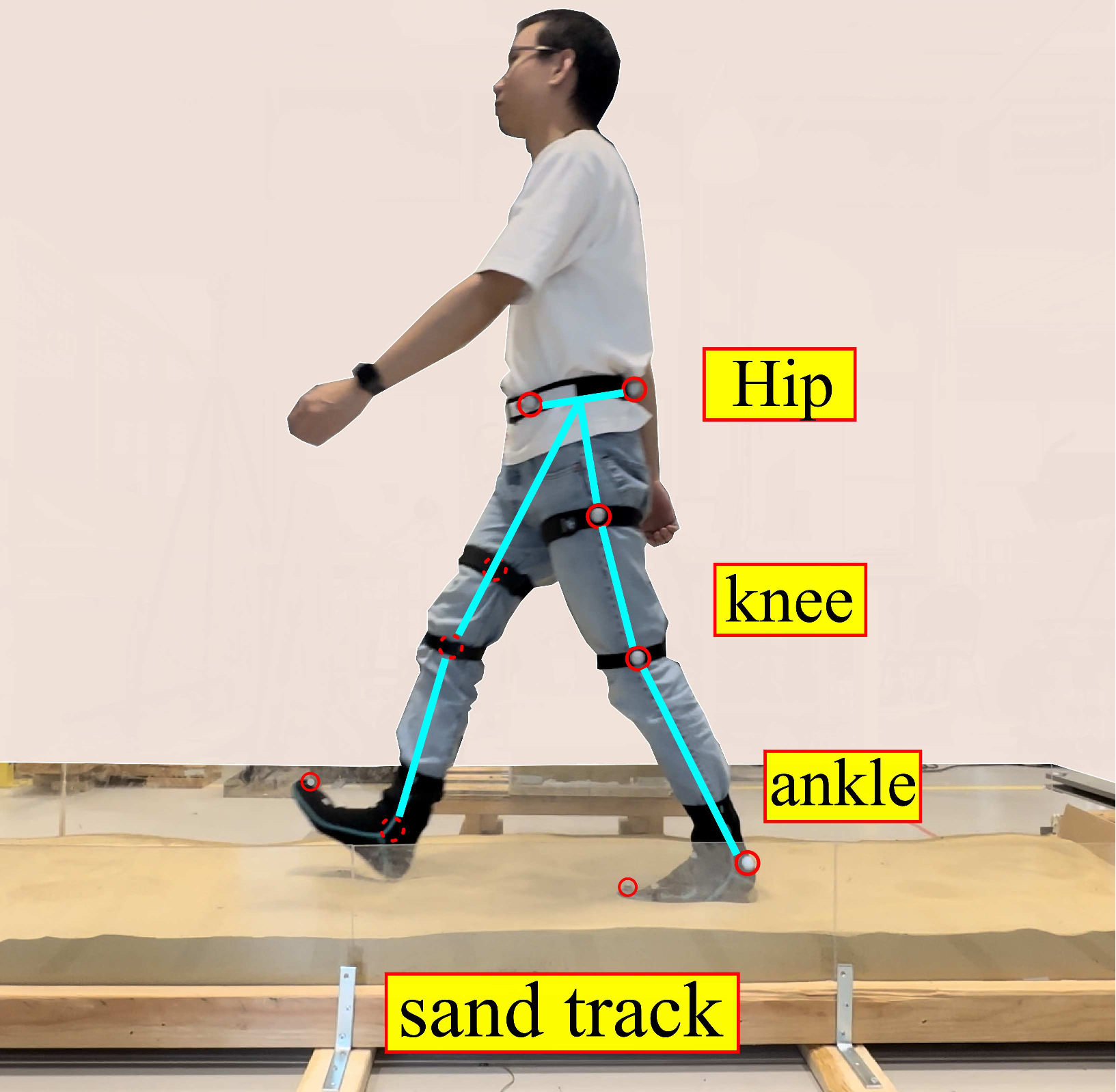}}
\subfigure[]{
	\label{fig:experimentSetup:c}
	\includegraphics[height = 1.75in]{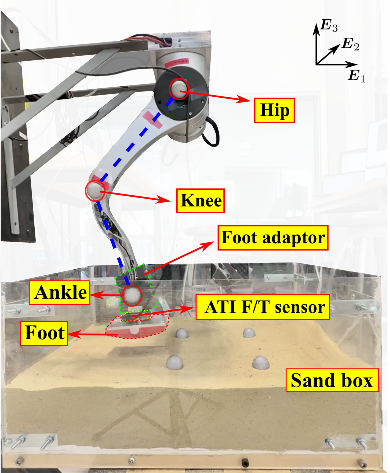}}
\subfigure[]{
	\label{fig:experimentSetup:d}
	\includegraphics[height = 1.8in]{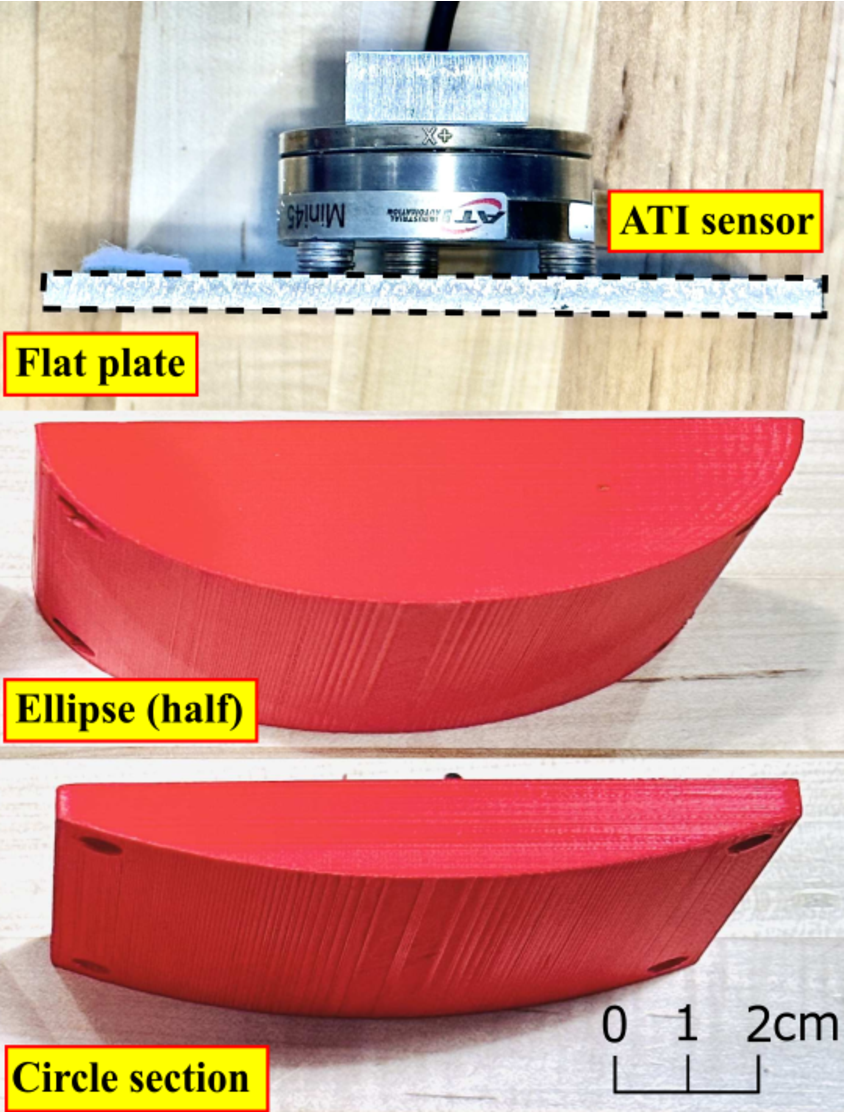}}
  \caption{The experimental setups. (a) RFT calibrations including vertical penetration tests for scale factor $\zeta$ and horizontal penetration tests for $f_1$ and $f_{23}$. (b) Humanoid walking gait capture using Vicon system. (c) Bipedal foot-terrain intrusions using a robotic leg. (d) Two selected non-flat foot models.}
 \label{fig:experimentSetup}
\vspace{-2mm}
\end{figure*}

\subsection{Force Model Experiments}

We conducted experiments to estimate the RFT model parameters, such as $\zeta$, $f_1$, $f_{23}$, and $\lambda_h$. Fig.~\ref{fig:experimentSetup:a} shows the overall experimental setup. One granular box ($51 \times 38 \times 25$~cm) was built and filled with fine sands (particle size around $0.2\backsim0.5$~mm) with a depth of $20$~cm. A robotic manipulator (Kinova Jaco) was used to control the intrusion motion for parameter estimation. A small metal plate ($35 \times 40 \times 1.3$~mm) was fabricated and mounted at the end-effector. A three-directional force/torque sensor (ATI Mini45) was mounted to measure the resistive forces during the intrusion. Optical markers were also attached to capture the real intrusion motion of the small plate.

Vertical penetration tests were conducted to obtain and estimate parameter $\zeta$. In order to estimate and obtain  two scaling factors $f_1$ and $f_{23}$, a flat plate was used for a sets of partial intrusion tests as shown in Fig.~\ref{fig:experimentSetup}(a). The plate was oriented vertically with an initial intrusion depth $z_0=10$~mm while the motion was initiated in the horizontal direction, i.e.,  $\beta=90^\circ$ and $\gamma = 0^\circ$. The plate orientation $\psi$ was then changed with an increment of $7.5^\circ$ and maintained the intrusion velocity $\|\bs{v}\|= 2$~cm/s. The normal force $F_{23}(\psi)$ and tangential force $F_{1}(\psi)$ were recorded. Two scaling factors were then calculated as
\begin{equation*}
  f_1 = \frac{F_1(\psi)}{F_1(0^{\circ})},\quad f_{23} = \frac{F_{23}(\psi)}{F_{23}(90^{\circ})}.
\end{equation*}
To keep the factor values within interval $[0,1]$, sigmoid functions were used, namely, $f_1 = \frac{a_{1}}{a_{2} + a_{3} \mbox{exp}(a_{4} \s_{\psi} + a_{5})}$, $f_{23} = \frac{b_{1}}{b_{2} + b_{3} \mbox{exp}(b_{4} \c_{\psi} + b_{5})}$, where $\s_{\psi} = \sin\psi$, $\c_{\psi} = \cos\psi$ and $a_i$, $b_i$, $i=1,...,5$, were fitting parameters whose values were determined by experimental data.

A set of partial intrusion tests using the flat plate were designed to estimate and obtain the values of the scaling factor $\lambda_h$. The flat plate was vertically inserted into substrates with a certain depth, $z_0$, and the motion was controlled and maintained in the horizontal direction, namely, $\beta=90^\circ$, $\gamma = 0^\circ$, and $\psi=90^\circ$. Different magnitudes of the intrusion velocity were tested and corresponding drag forces were recorded. By using~\eqref{eqn:modified_dF23}, parameter values for $\lambda_h$ were determined by the least-square method matching the experimental data. In experiments, $z_0= 1$~cm, $w=7$~cm, and $\phi_s=35^{\circ}$ were used.
\vspace{-1mm}

\subsection{Bipedal Robotic Walker Experiments}

To use a bipedal robotic walker in experiments, we first conducted human walking experiments on sand to obtain the gait. Fig.~\ref{fig:experimentSetup:b} shows an example of human subject experiments on sand. Human walking gaits were captured by the optical motion capture system (10 Bonita cameras from Vicon Ltd.) The hip, knee, and ankle positions were marked and relative positions of the knee and ankle with respect to the hip were calculated.  The hip and knee angles defined in Fig.~\ref{fig:robotWalker} were extracted. One complete walking gait was considered to start when the stance foot touched down the granular terrain and end as the same foot re-contacted the terrain. During the stance phase, the foot almost anchored in the soft terrain, and hip (body) moved forward, namely, if the hip (body) was fixed, the foot pushed the substrates downward and backward to obtain the supporting and thrust forces. We mainly consider the leg relative motion with respect to the hip to reconstruct the walking gait by pushing the foot in the substrates with the sand box fixed on the platform.

Fig.~\ref{fig:experimentSetup:c} illustrates the bipedal robotic foot intrusion experiments. The hip and knee joints were powered by brushless motors. A 3D-printed foot was installed at the ankle using a custom-built foot adaptor. Due to the hardware limitation, no ankle actuator was used for foot intrusion experiments. A three-directional force/torque sensor (ATI Mini45) was sandwiched between the adaptor and the foot. Optical markers were attached to the hip, knee, and ankle positions to capture the joint motion. Four additional markers were inserted inside sand to indicate the free surface. Three types of foot shapes reported to represent human feet were designed and tested~\cite{rodman2021developing}: flat, circular, and elliptical feet as shown in Fig.~\ref{fig:experimentSetup:d}. All three foot models shared the same length and width (i.e., $l\times w= 11 \times 7$~cm). The filled sand height was chosen such that the maximum magnitude of the vertical resistive force during the intrusion was approximately the value of the half weight of the robot leg assembly (around $13$~kg). The trajectory (i.e., path and velocity) of the leg intrusion were captured and differentiated by controlling the walking gait period $T_g$. Motion information was regarded as input for the RFT calculation.

\section{Experimental Results}
\label{Sec:result}

\begin{figure*}[h!]
	\centering
	\subfigure[]{
		\label{fig:Calibrations_VI}
		\includegraphics[height=1.5in]{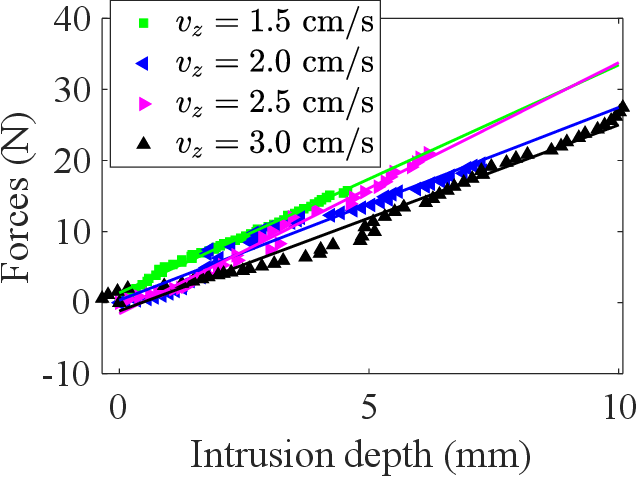}}
\hspace{1mm}
	\subfigure[]{
		\label{fig:Calibrations_HI}
\includegraphics[height=1.5in,width=2.10in]{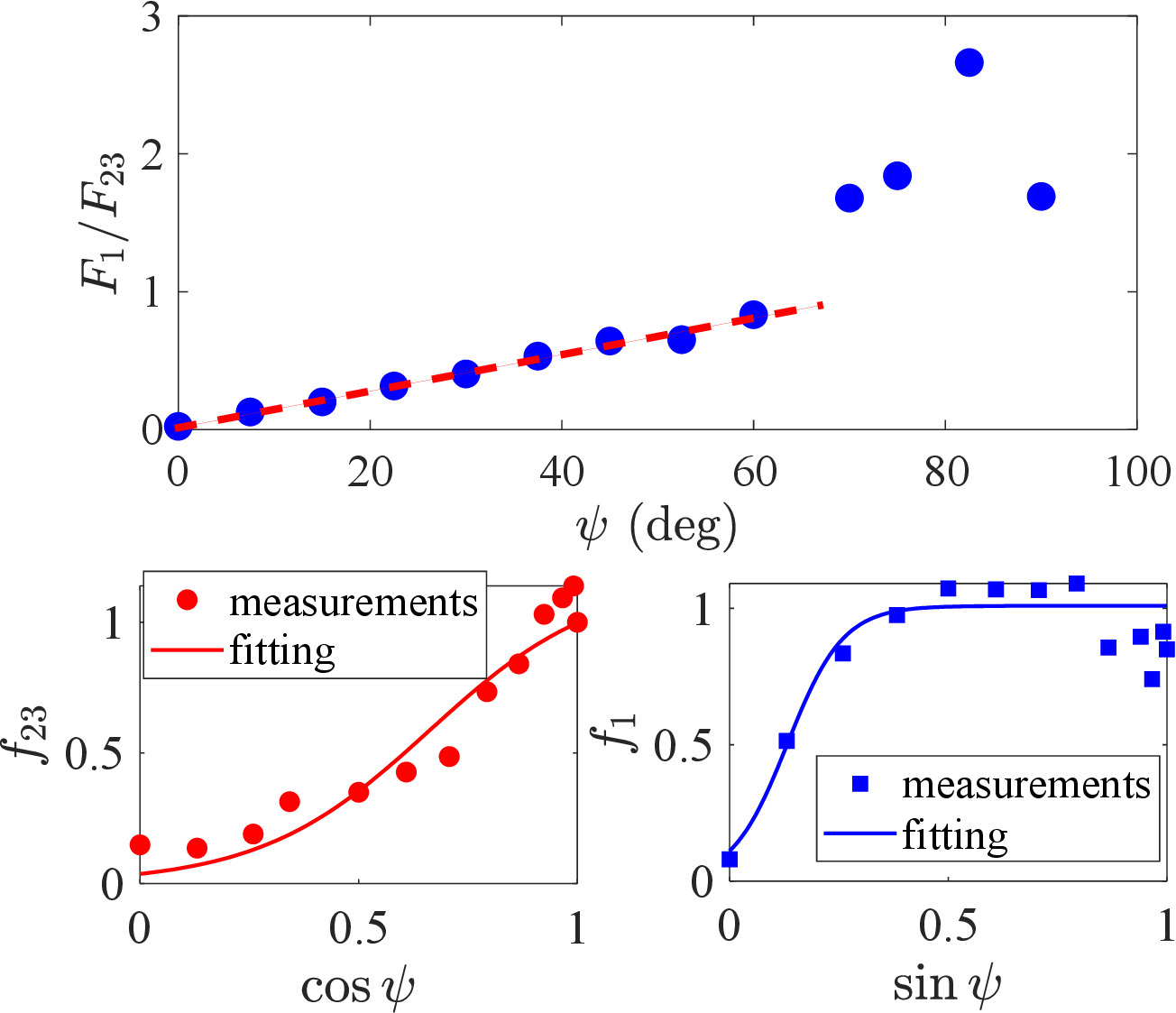}}
\hspace{2mm}
    \subfigure[]{
		\label{fig:Calibrations_Correction}
\includegraphics[height=1.5in]{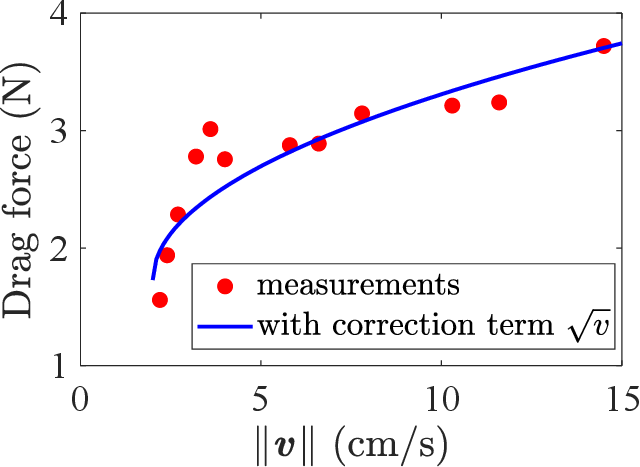}}
\vspace{-2mm}
\caption{Plate calibration results. (a) Vertical penetration resistive force versus intrusion depth. (b) Tangential and normal force relationship. $f_1$ and $f_{23}$ were fitted by sigmoid functions. (c) Horizontal penetration results under different velocities.}\label{fig:Calibrations}
\end{figure*}

\begin{figure*}[t!]
  \hspace{-2mm}
\subfigure[]{
\label{fig:legForceValidation:a}
 \includegraphics[height=2.3in,width=1.45in]{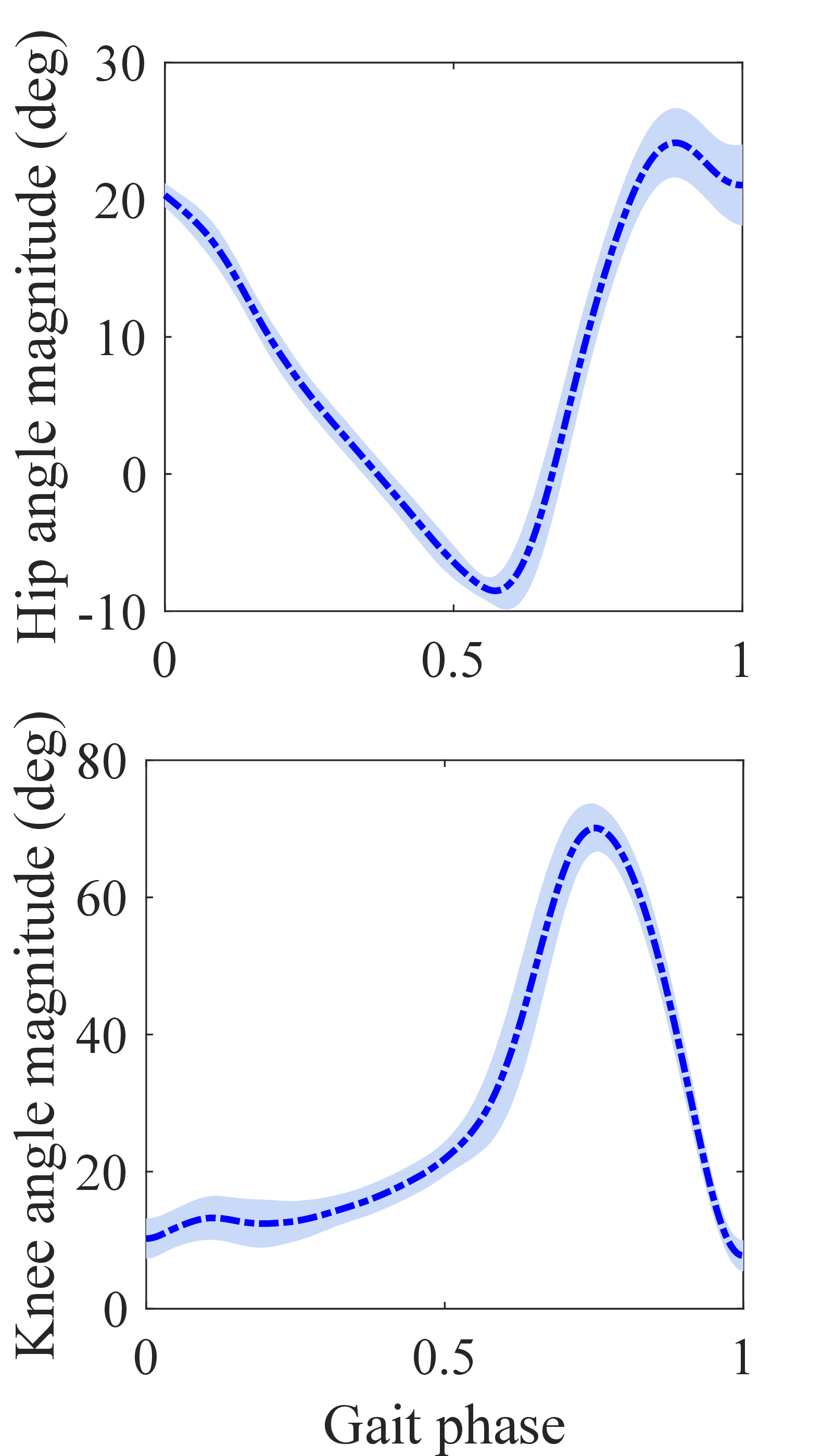}}
\hspace{-4mm}
\subfigure[]{
	\label{fig:legForceValidation:b}
	\includegraphics[height=2.25in,width=4in]{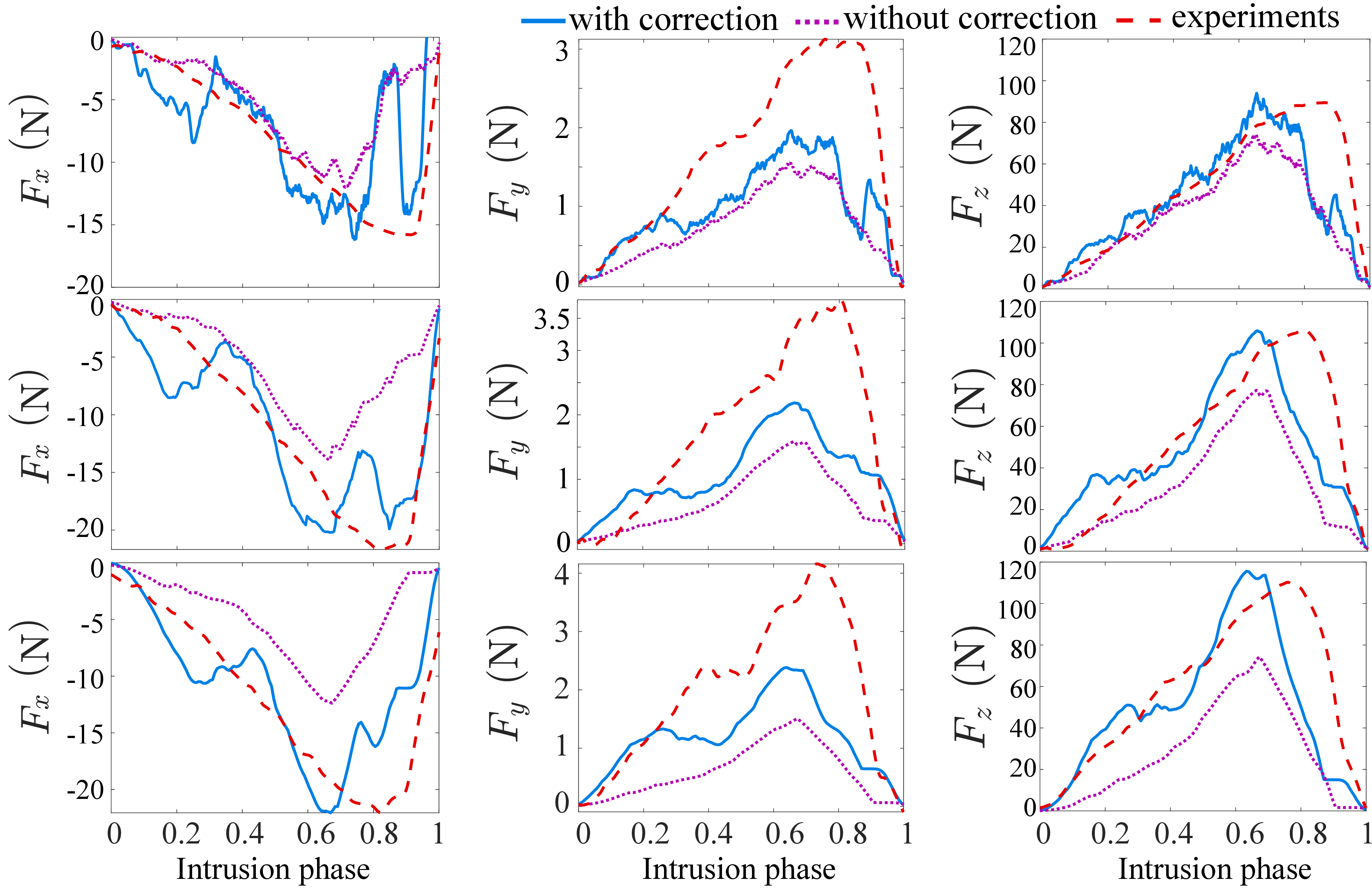}}
\hspace{-2mm}
\subfigure[]{
	\label{fig:legForceValidation:c}
	\includegraphics[height=2.2in]{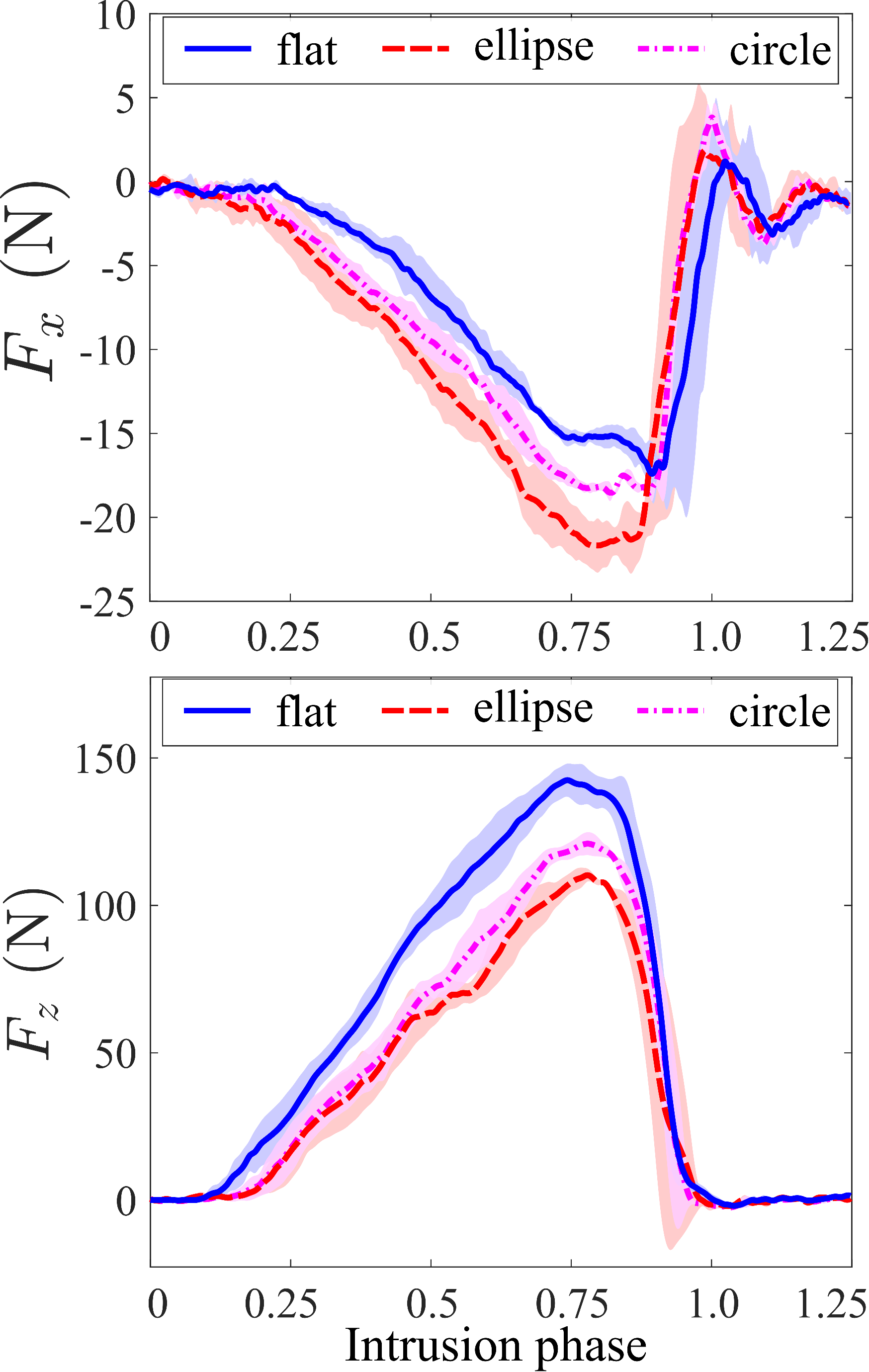}}
\vspace{-2mm}
  \caption{Dynamic RFT for foot-terrain intrusion results. (a) Human walking gait re-construction. (b) Forces of the robotic foot (ellipse) in sands. Blue solid lines present simulation results using the proposed method with the structural correction term. The first, second and third columns are for $F_x$, $F_y$, and $F_z$, respectively. The top, middle, and bottom rows show the slow ($T_g=13.5$~s), medium ($T_g=4.5$~s) and fast ($T_g=2.3$~s) gait speed intrusions, respectively. (c) The forces of the robotic foot with three shapes.}
  \label{fig:legForceValidation}
\vspace{-3mm}
\end{figure*}

\subsection{Force Model Validation}

Fig.~\ref{fig:Calibrations} shows the plate calibration results for the RFT model. Fig.~\ref{fig:Calibrations_VI} shows corresponding force results for vertical penetration under different velocities. The results indicated that vertical intrusion velocity impacted insignificantly. We used the average linear fitting curve to compute $\zeta=2.06$ for sand, which is consistent to the reported value in~\cite{HuangRAL2022}. As shown in Fig.~\ref{fig:Calibrations_HI}, the ratio $F_{1}/F_{23}$ increased almost linearly in the certain region (i.e., $\psi\in [0, 60^\circ]$). To quantify two factors $f_1$ and $f_{23}$, the values for the sigmoid function parameters were determined as: $a_1=1.15$, $a_2=1.14$, $a_3=1.82$, $a_4=-15.78$, $a_5=1.62$, $b_1=1.99$, $b_2=1.70$, $b_3=2.49$, $b_4=-5.17$, and $b_5=3.04$. Fig.~\ref{fig:Calibrations_Correction} shows the drag force changes as function of $\sqrt{\|\boldsymbol{v}\|}$. The fitting curve with scaling factor $\lambda_h=1.93$ matched well with the experimental data. For scaling factor $\lambda_v$ for inertial effect, we used the value $\lambda_v = 1.1$, which is similar to other reported studies~\cite{HuangRAL2022,AgarwalSA2021}. Note that all the parameters above are granular material-dependent and after calibrations, they can be used for various kinematic conditions.

\begin{figure*}[h!]
	\centering
	\includegraphics[height=2.9in,width=6.9in]{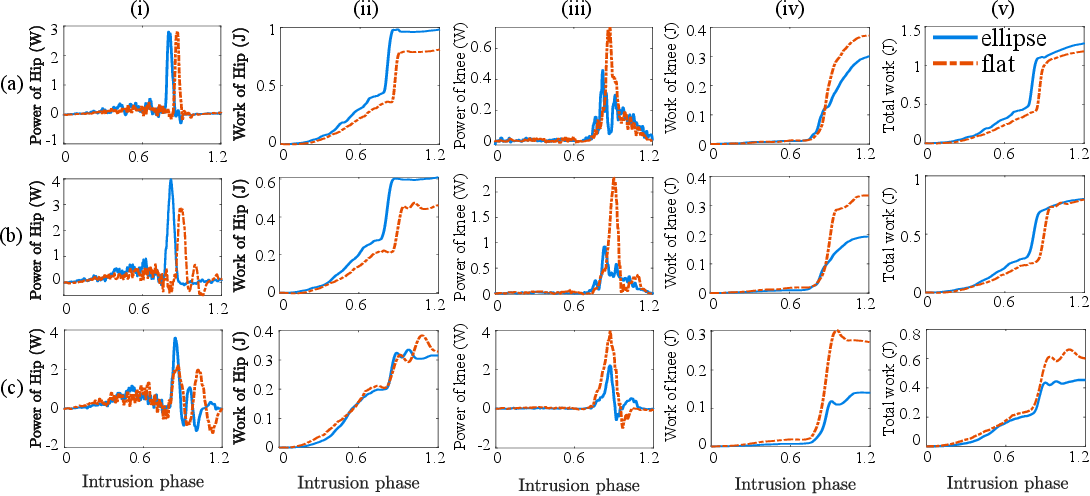}
\vspace{-1mm}
	\caption{The power and work performed by joint actuators with different gait speeds and different foot shapes. Rows (a)-(c) represent results of the slow ($T_g=13.5$~s), medium ($T_g=4.5$~s), and fast ($T_g=2.3$~s) gait speeds, respectively. Columns (i)-(ii) and (iii)-(iv) are instantaneous power and cumulative work at the hip and knee joints, respectively. Column (v) is the total resultant work.}
\label{fig:shapeOnEnergy}
\vspace{-1mm}
\end{figure*}

\subsection{Robotic Foot-Terrain Interactions}

Fig.~\ref{fig:legForceValidation:a} shows the hip and knee joint angles (mean and standard deviation) during human walking gait on sand. We took the average value for both joints as the tracking reference profile for robotic leg control. To consider the gait velocity influence, the robot leg was controlled to finish a complete gait within a certain period. From normal human walking, the forward speed was $1.2$~m/s associated with a gait period $1.1$~s according to the captured data. We used a small-size robot and controlled gait periods  $T_g=13.5$, $4.5$, and $2.3$~s to provide equivalent forward velocities $0.05$, $0.15$, and $0.28$~m/s, respectively. For convenience, we use the gait period $T_g$ to indicate how fast the intrusion was inside the granular media. Motion data were used in the simulations for validations of the proposed resistive force calculation approach.

For demonstration of foot-terrain interaction force validation, we illustrated the results using the elliptical shape foot as an example. Fig.~\ref{fig:legForceValidation:b} shows the 3D force validation results. The forces were presented as a function of the intrusion phase which began as the vertical resistive force was greater than zero and ended when the value of $F_z$ returned to zero. The purple dotted lines in the figure represent the simulation results using the conventional 3D-RFT method. The red dashed lines are the experimental results. As shown in Fig.~\ref{fig:legForceValidation:b}, the estimation of the drag force $F_x$ and the lifting force $F_z$ demonstrated the similar trend compared to the experimental results during the intrusion process (i.e., intrusion phase between $0$ and $0.7$). There are some differences for the lateral force $F_y$. The main reasons for such discrepancy are from several aspects. We used $\alpha_y = \alpha_x(0,0)$ and that might underestimate the tangential stress characteristics. Although the robotic leg was controlled in the sagittal plane, there might be induced torque at the foot in other directions, which would result in additional longitudinal and lateral forces.

Table~\ref{tab:RMSE} further lists the comparison force estimation results with and without the structural correction term in the proposed REF model under different walking speeds. As the gait speed increased, the RFT method with structural correction term can still estimate resistive forces closely, while the conventional RFT provided significant errors in terms of the force magnitude. These results confirm that the proposed enhancement of the RFT method provides an accurate estimation of the resistive forces for bipedal walking on granular terrains.

\renewcommand{\arraystretch}{1.3}
\setlength{\tabcolsep}{0.03in}
\begin{table}[t!]
\centering
\caption{RMSE of the RFT models under different gait speeds with and without correction term~\eqref{eqn:effectiveDepth}.}
\label{tab:RMSE}
\resizebox{\columnwidth}{!}{
\begin{tabular}{ccccccc}
\hline
       & \multicolumn{2}{c}{$F_x$}           & \multicolumn{2}{c}{$F_y$}           & \multicolumn{2}{c}{$F_z$}           \\ \cline{2-7}
       & with & w/o  & with & w/o  & with & w/o \\ \hline
slow   & $5.8\pm4.7$ & $7.5 \pm 5.3$ & $0.9\pm 0.6$ & $1.1\pm 0.7$ & $21.4\pm 19.5$ & $25.8 \pm 18.9$  \\
medium & $3.6\pm 3.3$ & $7.8\pm5.3$ & $1.0\pm0.8$ & $1.3\pm 0.8$ & $26.6\pm25.4$ & $35.8\pm26.4$  \\
high   & $3.8\pm2.7$ & $8.2\pm5.4$ & $1.1\pm0.8$ & $1.5\pm0.9$ & $22.8\pm20.4$ & $31.7\pm 19.6$  \\ \hline
\end{tabular}
}
\vspace{-3mm}
\end{table}

We implemented a set of foot intrusion experiments using different foot shapes and different intrusion speeds. Fig.~\ref{fig:legForceValidation:c} shows the resultant resistive forces on sand with three different shape feet. The walking gait period was $T_g=4.5$~s. The foot shape influenced the interaction forces and all three-directional forces experienced similar trend. Given the same gait, the elliptical shape foot generated the most drag force but the least lifting force during the intrusion. On the other hand, the flat foot provided the least drag force because the resistive force under this situation was mainly the tangential force, i.e., $f_1$ became a dominant term for the RFT calculation in~\eqref{eqn:modified_3D RFT}.

Fig.~\ref{fig:shapeOnEnergy} shows the instantaneous power and cumulative work performed by the joint actuation for different foot shapes. We only show the results for the elliptical and flat shape feet because the results of the circular shape foot are similar to the elliptical one. Therefore, we used elliptical shape to represent curved foot type for a concise result representation. The results were compared under three gait speeds. It is interesting to find that the different gait speed would lead to the different foot shape preference in work performed and energy consumption of the robot. The peak of the power generated by the hip joint occurred when the intrusion happened and the profile trend did not change significantly for elliptical and flat feet under three gait speed conditions. Compared with the flat foot, the elliptical foot required more hip actuator power to overcome the resistive forces and to maintain the gait motion, which naturally resulted in more work and energy consumption. Considering that the duration of the intrusion became shorter as the gait speed increased, the cumulative work performed by the hip actuator decreased for both elliptical and flat foot. In terms of the knee joint, the power increased under a fast gait speed. The elliptical shape foot showed a lower level of power requirement. Therefore, the cumulative work done by the knee joint using the elliptical foot was significantly less than that of the flat foot. Based on the analysis of the experimental results, the elliptical shape foot was suitable for bipedal walking with a high gait speed, which required less work than the flat foot to compensate resistance forces on the soft terrain. For slow walking, the flat foot however enabled bipedal walkers to save energy and instead became a preferable option.

\vspace{-1mm}
\section{Conclusion}
\label{Sec:conc}

This paper proposed an enhanced resistive force model by introducing an additional intrusion depth correction term for bipedal robot walking on granular terrains. To validate the force model for walking applications, robotic foot intrusion test kit were built and three 3D-printed feet with different shapes. The reaction forces such that drag force and lifting force, and power/energy were investigated considering the gait speed and foot shape influence. The elliptical shape foot showed a lower level of power requirement for high-speed-gait walking while the flat foot design was suitable for slow-motion walking instead. As an ongoing effort, we plan to explore the optimal walking gait and foot shape for bipedal walking. Another future research direction is to use the proposed force model for robotic control.

\bibliographystyle{IEEEtran}
\bibliography{Chen_ICRA2024_Ref}
\end{document}